\documentclass[runningheads]{llncs}
\usepackage[utf8]{inputenc}
\usepackage[T1]{fontenc}
\usepackage{amsmath,amssymb}
\usepackage{booktabs}
\usepackage{hyperref}
\usepackage{xcolor}
\usepackage{todonotes}
\usepackage{float}

\usepackage{color}
\usepackage{isabelle,isabellesym}

\definecolor{IsaBlue}{cmyk}{1.0,0.33,0,.4}
\definecolor{IsaGreen}{cmyk}{1.0,0,1.0,.2}
\definecolor{IsaRed}{cmyk}{0,0.69,0.69,0}
\definecolor{IsaDarkRed}{cmyk}{0,1.0,1.0,0.15}

\newcommand{\isakwmaj}[1]{\textcolor{IsaBlue}{\textbf{\texttt{#1}}}}
\newcommand{\isakwmin}[1]{\textcolor{IsaGreen}{\textbf{\texttt{#1}}}}

\newcommand{\isacmt}[1]{\textcolor{IsaDarkRed}{\texttt{#1}}}

\isabellestylett
\renewcommand{\isastyle}{\isastyleminor}

\newcommand{\isakw}[1]{{\bfseries\ttfamily\def\isachardot{.}\def\isacharunderscore{\isacharunderscorekeyword}%
\def\isacharbraceleft{\{}\def\isacharbraceright{\}}#1}}

\renewcommand{\isakeyword}[1]{\textcolor{IsaGreen}{\isakw{#1}}}

\newcommand{\ismark}[1]{\isakwmaj{#1}}
\newcommand{\ismin}[1]{\isakwmin{#1}}

\newcommand{\isstr}[1]{\texttt{#1}}
\renewcommand{\isa}[1]{\texttt{#1}}

\makeatletter
\renewcommand\section{\@startsection{section}{1}{\z@}%
                       {-12\p@ \@plus -4\p@ \@minus -4\p@}%
                       {8\p@ \@plus 4\p@ \@minus 4\p@}%
                       {\normalfont\large\bfseries\boldmath
                        \rightskip=\z@ \@plus 8em\pretolerance=10000 }}
\makeatother

\makeatletter
\renewcommand{\paragraph}[1]{%
  \@startsection{paragraph}{4}{0pt}{3pt}{0pt}{\itshape}%
  {#1}\hspace{1pt}
}
\makeatother

\begin{document}

\title{Munkres' General Topology Autoformalized\\ in Isabelle/HOL}

\author{Dustin Bryant\inst{1} \and
Jonathan Juli\'an Huerta y Munive\inst{2} \and
Cezary Kaliszyk\inst{3} \and
Josef Urban\inst{4}}

\titlerunning{Munkres' Topology Autoformalized in Isabelle/HOL}
\authorrunning{D. Bryant, J.\,J. Huerta y Munive, C. Kaliszyk, J. Urban}

\institute{Independent, \email{brydustin@gmail.com} \and
Aalborg University, \email{jjhymuni@cs.aau.dk} \and
University of Melbourne, \email{ckaliszyk@unimelb.edu.au} \and
AI4REASON and University of Gothenburg / Chalmers University
\email{josef.urban@gmail.com}}

\maketitle

\begin{abstract}
We describe an experiment in LLM-assisted autoformalization that produced
over 85,000 lines of Isabelle/HOL code covering all 39~sections of Munkres'
\emph{Topology} (general topology, Chapters 2--8), from topological spaces through dimension theory.
The LLM-based coding agents (initially ChatGPT~5.2 and then Claude~Opus~4.6) used 24 active days for that.
The formalization is \emph{complete}: all 806 formal results are fully proved with zero \isa{sorry}'s.
Proved results include the Tychonoff theorem, the Baire category
theorem, the Nagata--Smirnov and Smirnov metrization theorems,
the Stone--\v{C}ech compactification, Ascoli's theorem, the space-filling curve, and others.

The methodology is based on a ``sorry-first'' declarative proof workflow combined with
bulk use of \isa{sledgehammer} - two of Isabelle major strengths. This leads to relatively fast autoformalization progress.
We analyze the resulting formalization in detail,
analyze the human--LLM interaction patterns from the session log,
and briefly compare with related autoformalization efforts in Megalodon,
HOL~Light, and Naproche.
The results indicate that LLM-assisted formalization of standard
mathematical textbooks in Isabelle/HOL is quite feasible, cheap and fast, even if some human supervision is useful.
\end{abstract}

\section{Introduction}
\label{sec:intro}

Formalization of mathematics--- translation of informal textbook proofs
into machine-checkable code---has typically required substantial
manual effort.
Major formal proof libraries such as Isabelle's Analysis library~\cite{holanalysis}, Lean's Mathlib,
and the Mizar Mathematical Library~\cite{MML2017} represent years or decades of cumulative work by
dedicated communities.
The possibility that large language models might significantly
accelerate this process has therefore attracted considerable interest.

In January 2026, the fourth author reported an experiment in which
approximately 130,000 lines of formal topology were produced in roughly
two weeks, formalizing large portions of Munkres'
\emph{Topology}~\cite{munkres} in the Megalodon proof
system~\cite{urban2026megalodon} using ChatGPT~5.2 via the Codex CLI.
Despite being conducted in a relatively lightweight system with limited
prior training data, the experiment demonstrated that large-scale
autoformalization is feasible in practice.

The present paper describes a related experiment:\footnote{The code is at \url{https://github.com/JUrban/isa_top_autoform1} .} the autoformalization
of the same textbook in {Isabelle/HOL}~\cite{nipkow2002isabelle},
one of the field's most mature and well-supported proof assistants.
In this setting, the LLM has access to Isabelle's extensive library
(\isa{Complex\_Main}), powerful proof automation
(\isa{sledgehammer}~\cite{blanchette2016hammering}, \isa{blast},
\isa{auto}, etc.), and the structured (declarative) Isar proof language.
These features provide a substantially richer environment than in the
Megalodon experiment.
A natural question is whether such infrastructure significantly reduces
the difficulty and speeds up the large-scale formalization.
In practice, both effects are observed: the available automation and
library support simplify and speed up some aspects of the development, while
on the other hand, the more complex setting requires more involved oversight of the available tools.

\paragraph{Main results.}
Over 24 active days (March 2 -- April 3, 2026), the project produced
85,472 lines of Isabelle/HOL across four chapter files, comprising
199 definitions and 806 lemmas, theorems, and corollaries covering all
39 sections (\S\S12--50) of point-set topology in Munkres' book.
The formalization is \textbf{complete}: all results are fully proved
with zero \isa{sorry}'s.
Chapter~1 (Set Theory and Logic) is covered by Isabelle's \isa{Complex\_Main}.
Among the proved results are the Tychonoff theorem,
the Baire category theorem (both parts), the Nagata--Smirnov and Smirnov
metrization theorems, the Stone--\v{C}ech compactification,
both the classical and general versions of Ascoli's theorem
(Theorems~45.4 and~47.1---the latter depending on Tychonoff),
Theorem~48.5 (pointwise limits on Baire spaces),
the space-filling curve (Theorem~44.1),
and the existence of a nowhere-differentiable function (Theorem~49.1).
 
\paragraph{Methodology.}
The experiment used two LLMs in sequence---ChatGPT~5.2 for bulk
formalization and Claude~Opus~4.6 for proof development---both driven
by an automated tmux-based prompting loop.
The central methodological idea is a ``sorry-first'' workflow:
proof skeletons are written with placeholders, which are subsequently
filled using automated reasoning tools such as \isa{sledgehammer}.
This separates proof \emph{structure} from proof \emph{search}.

\paragraph{Contributions.}
We produced: (1)~a \emph{complete} autoformalization of Munkres' general topology in a mainstream proof
assistant---all 806 results fully proved with zero \isa{sorry}'s. While we evolved the prompt and the methods, no part of the code has been written by a human (this is an intentional design choice of this experiment).\footnote{It is obviously possible to include human modifications and this will likely be done in various ways with this material in the future. However, direct and continued human modification usually makes the scientific results of autoformalization experiments and evaluations much less clear.}
(2)~A methodology for the long-running LLM--Isabelle interaction largely based on the
sorry-first/sledgehammer workflow.
(3)~Analysis of the session log,
revealing patterns of human--LLM interaction.
(4)~A comparison with several related autoformalization efforts on
the same source material. (5)~A qualitative evaluation of the
mathematical adequacy of the formalization.

\section{Setup}
\label{sec:setup}

\subsection{Source Material}

The formalization target is a LaTeX representation of Munkres'
\emph{Topology}~\cite{munkres} (\isa{top1.tex}, 7,956 lines), covering
Chapters~2--8 (Sections~12--50):
topological spaces and continuous functions (\S\S12--22),
connectedness and compactness (\S\S23--29),
countability and separation axioms (\S\S30--36),
the Tychonoff theorem (\S37--38),
metrization and paracompactness (\S\S39--42),
complete metric spaces and function spaces (\S\S43--46),
and Baire spaces through dimension theory (\S\S47--50).
The source contains 76 definitions, 109 theorems, 34 lemmas,
13 corollaries, and 115 examples;
exercises are excluded from the formalization.

\subsection{Isabelle/HOL}

Isabelle~\cite{isabelle} is a generic logical framework whose
principal instantiation, Isabelle/HOL~\cite{nipkow2002isabelle},
provides higher-order logic with
a large standard library.
Our formalization only imports \isa{Complex\_Main}, giving it access to the reals and
complex numbers, topological spaces (as a type class), metric spaces,
continuity, compactness, Heine--Borel, measure theory, and extensive
algebraic infrastructure.
This is a considerable advantage over the Megalodon experiment, where the
library consists of Brown's set-theoretic foundations with reals constructed
via surreal numbers~\cite{brown2025hammering}, but without the
extensive analytical 
 infrastructure that Isabelle provides. %

The proof automation is correspondingly richer.
Beyond \isa{sledgehammer}~\cite{blanchette2016hammering}---which connects
to external ATPs (Vampire, CVC5, Z3) with premise selection and proof
reconstruction---Isabelle offers \isa{blast} (classical tableaux),
\isa{auto}/\isa{simp} (rewriting + classical reasoning),
\isa{linarith}/\isa{presburger} (arithmetic decision procedures),
and
\isa{meson}/\isa{metis} (first-order provers).
Megalodon has the \isa{aby} hammer~\cite{brown2025hammering}, which
exports to TH0 TPTP~\cite{th0} and has good ATP performance but lacks premise
selection and proof reconstruction---a critical limitation for the
LLM workflow.

\subsection{LLM Agents}

The project used two LLMs in sequence:

\noindent\textbf{Phase~1} (March 2--13): {ChatGPT~5.2} with high
reasoning, via OpenAI's Codex CLI---the same model and interface as the
Megalodon experiment.
This phase (running continuously on a server) produced 268 commits (``topology 0001'' through ``topology
0282''), formalizing all definitions and theorem statements with proofs
often deferred via \isa{sorry}.

\noindent\textbf{Phase~2} (March 21 -- April 3): {Claude~Opus~4.6}
via Claude Code (Anthropic's CLI agent---running with pauses on the fourth author's notebook), focused on systematic proof
development.
This phase produced 884 commits with detailed theorem-by-theorem
progress, reducing the sorry count from 51 to \textbf{zero}.

Both agents ran in an automated tmux-based loop: a script monitors
the session and re-issues the prompt (``Read CLAUDE.md and follow
instructions'') whenever the agent finishes or stalls---the same
mechanism used in the Megalodon experiment.
The human occasionally intervened for focused direction or to correct
misbehavior (see Section~\ref{sec:session}).

\subsection{Tooling Evolution}

As the project progressed, several infrastructure improvements wee developed.
A session-splitting optimization---caching Chapters~2--4 in a separate
precompiled heap image---cut incremental build times from 50--60\,s to
12--13\,s. Since our workflow is compilation based, this is critical for fast feedback and progress. 
Performance profiling revealed that slow \isa{metis} calls in
equality-free topology proofs could be replaced by \isa{meson}, often
yielding 10$\times$ speedups.
Custom Isabelle CLI tools were developed for the project:
\isa{eval\_at} (with per-line timing via the \isa{-t} flag)
and \isa{desorry} (for automated sorry elimination).
Together with Isabelle's built-in \isa{process\_theories} command
(which runs theory processing outside the session timeout),
these tools enabled the bulk sledgehammer workflow described in the
next section.

\section{The Sorry-First Methodology}
\label{sec:method}

\begin{figure}[htbp]
  \centering
      \includegraphics[angle=90,width=0.7\textheight]{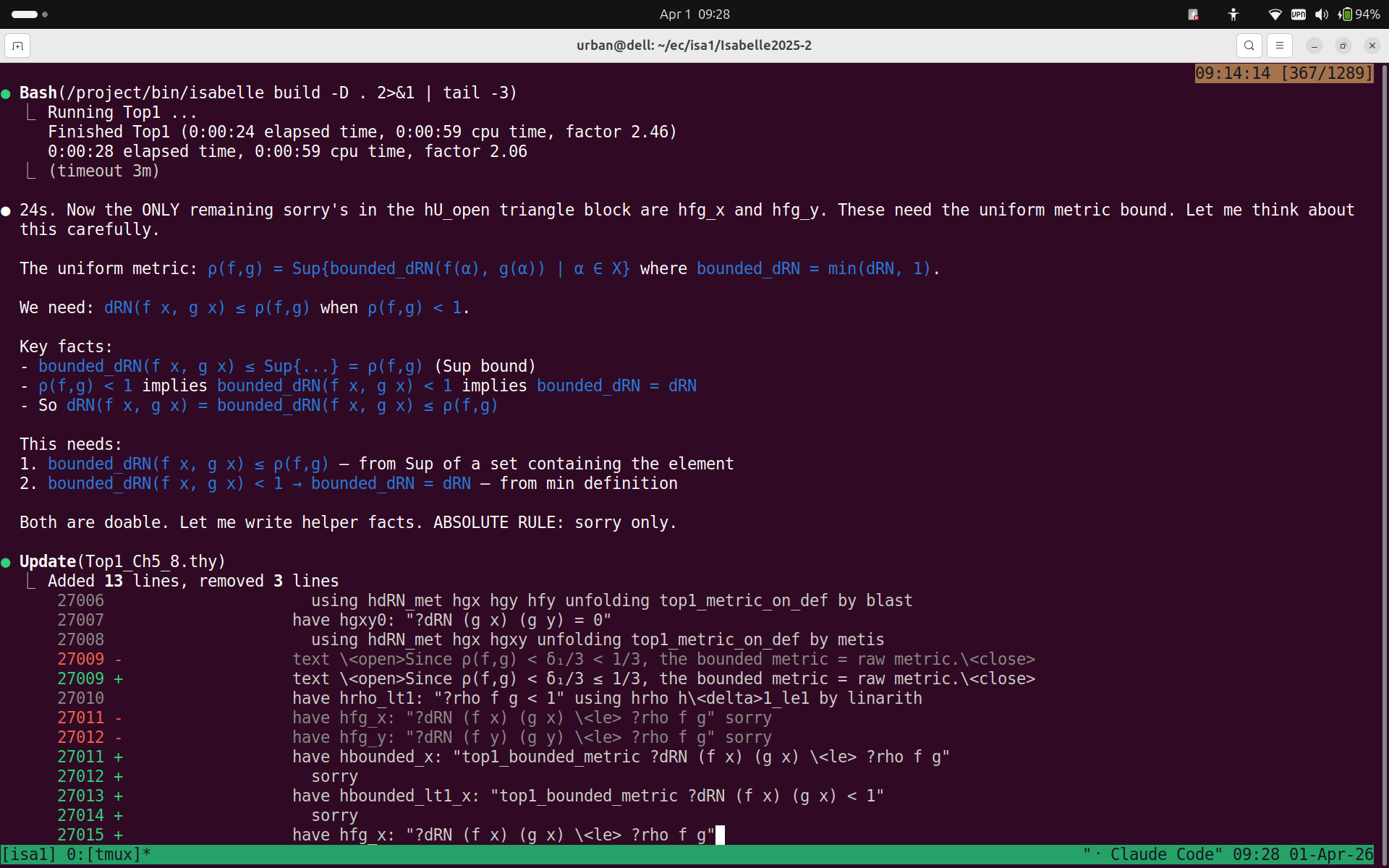}
    \caption{Example of the LLM's strategic thinking and use of the sorry-based method.}
    \label{fig:screen1}
\end{figure}

The central methodological insight, hard-won through repeated encounters
with Isabelle's capacity for combinatorial explosion, is the
\emph{sorry-first workflow} (see also Fig~\ref{fig:screen1}):

\begin{enumerate}
\item Write the proof skeleton with \isa{sorry} at \emph{every} step.
      No other tactic is permitted in new code---``not \isa{by fast},
      not \isa{by linarith}, not \isa{by simp}, not \isa{by (rule ...)},
      not even \isa{done},'' as the final version of the instructions puts it,
      reflecting successive refinements of the instruction set.
\item Build to verify the skeleton is well-formed.
\item Annotate all sorry'd steps with \isa{sledgehammer [timeout = 10]}.
\item Run \isa{process\_theories} \emph{once} to collect all ATP
      suggestions in a single pass.
\item Substitute the fastest proof for each step.
\item When sledgehammer fails, decompose the step into finer \isa{have}
      blocks and repeat.
\item Use \isa{eval\_at -t} to profile and optimize slow tactics.
\end{enumerate}

This rule is the culmination of a rather painful evolutionary process.
The models have a natural %
tendency to write proof methods immediately.\footnote{This seems to be an artifact of the unconstrained LLM training on the large Isabelle corpora available online. It is interesting to note -- as an counterexample to the accepted LLM ``token prediction is all you need'' wisdom -- that an (L)LM trained on a modified corpus with the proof methods removed would be very useful and likely much better behaved for our approach.}
Unfortunately, in a context importing \isa{Complex\_Main}---which
contributes thousands of simplification rules---even an innocent-looking
\isa{by auto} can trigger minutes of fruitless search.
The instruction file (\isa{CLAUDE.md}) went through eight revisions,
each prompted by a specific failure mode (see Section~\ref{sec:session}).

One consequence of this workflow is that the LLM did not extend the
simplifier set in any way. Similarly, it did not add any other automation-related
attributes to any theorems (such as \isa{[intro]} or \isa{[elim]}).
All 8,160 \isa{unfolding} calls and the dominance of explicit
\isa{by blast} (8,879 uses) over \isa{by auto} (576 uses) reflect
this: each definition is manually unfolded at every use site,
rather than registered for automatic simplification.
This is decidedly non-standard Isabelle practice, but it is \emph{safe}:
it prevents the cascading simpset explosions that bedevil large
developments.

\section{The (Auto-)Formalization}
\label{sec:formalization}

\subsection{Design}

Unlike previous formalizations of topology that rely on type 
classes~\cite{typeclasses}, type definitions~\cite{holanalysis}, or 
locales~\cite{notsimplett}, the LLM chose (independently of the humans) to use 
\emph{explicit carrier sets}, e.g. a topology as a pair
$(X, \mathcal{T})$ with $\mathcal{T} \subseteq \mathcal{P}(X)$
is formalized as shown below.

\begin{quote}\isaspacing\isastyle
\ismark{definition} \isstr{is{\kern0pt}\_topology{\kern0pt}\_on} :: \texttt{'a set} $\Rightarrow$ \texttt{'a set set} $\Rightarrow$ \texttt{bool} \ismin{where}\\
\quad \isstr{``is{\kern0pt}\_topology{\kern0pt}\_on}\ X\ T $\longleftrightarrow$\\
\qquad $\{\}\ \in\ T\ \wedge\ X\ \in\ T\ \wedge\ (\forall U.\ U \subseteq T \longrightarrow \bigcup U \in T)\ \wedge$\\
\qquad $(\forall F.\ \mathrm{finite}\ F\ \wedge\ F \neq \{\}\ \wedge\ F \subseteq T \longrightarrow \bigcap F \in T)$\isstr{''}
\end{quote}

\noindent This mirrors Munkres' definitions and does not interoperate with
Isabelle's type-class-based topology (which allows only one topology per
type).  Names of lemmas and definitions in the formalization include a \isa{top1\_} prefix (again, a decision made  by the LLM).
They follow Munkres' pattern \isa{Theorem\_N\_M} and each is annotated with its source
location as follows:\\
\isacmt{(**~from~\S37~Theorem~37.3~(Tychonoff)~[top1.tex:5253]~**)}.

\subsection{Content}

\begin{table}[t]
\centering
\caption{Formalization content by chapter}
\label{tab:content}
\begin{tabular}{@{}lrrrrc@{}}
\toprule
File & Lines & Defs & Results & Sorry & Status \\
\midrule
Ch2 (\S\S12--22) & 20,751 & 78 & 189 & 0 & \checkmark \\
Ch3 (\S\S23--29) & 13,458 & 28 & 107 & 0 & \checkmark \\
Ch4 (\S\S30--36) & 18,908 & 18 & 123 & 0 & \checkmark \\
Ch5--8 (\S\S37--50) & 32,355 & 75 & 387 & 0 & \checkmark \\
\midrule
\textbf{Total} & \textbf{85,472} & \textbf{199} & \textbf{806} & \textbf{0} & \checkmark \\
\bottomrule
\end{tabular}
\end{table}

Table~\ref{tab:content} summarizes the formalization.
The 806 formal results include 195 directly corresponding to Munkres'
numbered results (covering all 156 non-exercise statements in the source,
with some split into parts) and 611 helper lemmas
(a helper-to-Munkres ratio of ${\approx}\,3.1{:}1$).

The initial Chapters~2--4 cover among other things:
the Urysohn lemma (Theorem~33.1), the Urysohn metrization theorem
(Theorem~34.1), the Tietze extension theorem (Theorem~35.1),
and the imbedding of manifolds (Theorem~36.2).
A comparison of such landmark proofs between different projects and systems is instructive, but one must be
careful about what is being counted.
In our formalization, the proofs are \emph{heavily factored} into
helper lemmas, so there are two natural measurements: the \emph{direct}
proof body of the theorem itself, and the entire \emph{section} including
all supporting infrastructure.
The Megalodon paper~\cite{urban2026megalodon} reports \emph{direct} proof
lengths (the theorem's own proof body, Table~3 therein);
Mulligan and Paulson~\cite{mulligan2026copilot} report a self-contained 275-line proof.

Table~\ref{tab:proofsize} gives both measurements for our formalization,
alongside the Megalodon and Mulligan--Paulson figures.
The direct proof comparisons are as follows: our Urysohn lemma
(Theorem~33.1) is 287 lines versus Megalodon's 2,964; our Tietze
extension (Theorem~35.1) is 324 lines versus Megalodon's 10,369---the
infamous ``Battle of the Tietze Hill,'' which required ${\sim}$20 hours
and 35 iterations in Megalodon~\cite{urban2026megalodon}.
These order-of-magnitude differences in direct proof length reflect
two factors: (i)~Isabelle's \isa{Complex\_Main} supplies analytical
prerequisites that Megalodon must build inline, and
(ii)~our factoring strategy moves a lot of content into helper lemmas
(the section totals of 3,126 and 3,597 lines are closer to the
Megalodon figures).
In effect, the same mathematical content is present but distributed
differently---concentrated in one large proof in Megalodon,
versus spread across 18--20 small lemmas in Isabelle.

The Mulligan--Paulson proof of the Urysohn metrization theorem
(275 lines) is much shorter than both our direct proof
(1,196 lines---a single structured proof block with 260 \isa{have}
steps that was not factored into helpers) and Megalodon's (2,174 lines).
This reflects their use of the type-class library where
\isa{metrizable\_space}, \isa{regular\_space}, and \isa{second\_countable}
are pre-existing concepts with substantial lemma infrastructure---our
formalization defines these from scratch with explicit carrier sets.
Our Theorem~34.1 is also a case where the sorry-first factoring
discipline was not consistently applied: the monolithic proof was
produced in Phase~1 by ChatGPT~5.2, before the sorry-first
methodology was adopted.

\begin{table}[t]
\centering
\caption{Proof sizes for landmark theorems}
\label{tab:proofsize}
\setlength{\tabcolsep}{4pt}
\begin{tabular}{@{}lrrrrr@{}}
\toprule
Theorem & Direct & Section & Helpers & Megal. & M--P \\
\midrule
Urysohn lemma (33.1) & 287 & 3,126 & 20 & 2,964 & --- \\
Urysohn metriz.\ (34.1) & 1,196 & 3,373 & 19 & 2,174 & 275 \\
Tietze extension (35.1) & 324 & 3,597 & 18 & 10,369 & --- \\
Tychonoff (37.3) & 198 & 1,394 & 6 & --- & --- \\
Nagata--Smirnov (40.3) & 968 & 2,455 & 38 & --- & --- \\
Smirnov metriz.\ (42.1) & 83 & 347 & 4 & --- & --- \\
Space-filling curve (44.1) & 325 & 1,919 & 44 & --- & --- \\
Ascoli general (47.1) & 372 & 1,912 & 29 & --- & --- \\
Baire category (48.2) & 62 & 2,907 & 21 & --- & --- \\
Thm~49.1 (nowhere-diff.) & 53 & 1,296 & 24 & --- & --- \\
Dim.\ imbedding (50.5) & 1,103 & 6,271 & 96 & --- & --- \\
\bottomrule
\end{tabular}
\smallskip

\noindent{\footnotesize
Direct = theorem's own proof body (comparable to Megalodon column).
Section = full section including all helper lemmas and definitions.
Megal.\ = Megalodon direct proof length (from~\cite{urban2026megalodon}, Table~3).
M--P = Mulligan--Paulson (self-contained, type-class-based).}
\end{table}

The contrast between ``Direct'' and ``Section'' columns reveals
the factoring strategy: the Baire category theorem's direct proof
is just 62 lines, but its section is 2,907 lines because the heavy
lifting is in the nested-ball construction helper
(\isa{top1\_baire\_nested\_ball\_helper}) and the compact-Hausdorff
case.  Similarly, Theorem~49.1's direct proof is 53 lines---it
merely assembles results from the 24 supporting lemmas that do the
real work.
This extreme decomposition is a direct consequence of the sorry-first
methodology: each helper is independently proved via sledgehammer,
and the main theorem becomes a short assembly of pre-verified pieces.

Table~\ref{tab:directproofs} gives a broader view, listing all
theorems with a direct proof of over 300 lines across the entire
formalization.
The longest direct proof is the Urysohn metrization theorem
(Theorem~34.1, 1,196 lines)---a monolithic Phase~1 proof
that predates the factoring discipline.
The second longest is the dimension-theory imbedding theorem
(Theorem~50.5, 1,103 lines), involving general-position arguments
and partitions of unity.
Among other Phase~2 proofs, the Nagata--Smirnov theorem (968 lines) and
Theorem~48.5 (833 lines) are the largest.
The median direct proof length for the 806 results is approximately
25 lines, reflecting the prevalence of short, focused lemmas
with just 2--5 \isa{have} steps each.

\begin{table}[t]
\centering
\caption{All direct proofs exceeding 300 lines}
\label{tab:directproofs}
\setlength{\tabcolsep}{4pt}
\begin{tabular}{@{}rlr|rlr@{}}
\toprule
\S & Theorem & Lines & \S & Theorem & Lines \\
\midrule
34 & Thm~34.1 (Urysohn met.) & 1,196 & 38 & Thm~38.2 (Stone--\v{C}ech) & 335 \\
50 & Thm~50.5 (dim.\ imbedding) & 1,103 & 44 & Thm~44.1 (space-filling) & 325 \\
21 & Lem.~21.4 (seq.\ closure) & 990 & 33 & Thm~33.2 (product) & 325 \\
40 & Thm~40.3 (Nagata--Sm.) & 968 & 29 & Thm~29.2 (1-pt compact.) & 325 \\
23 & Thm~23.6 (fin.\ prod.\ conn.) & 900 & 35 & Thm~35.1 (Tietze) & 324 \\
32 & Thm~32.4 & 839 & 45 & Thm~45.4 (Ascoli classical) & 319 \\
48 & Thm~48.5 (ptw.\ limits) & 833 & 25 & Thm~25.2 (components) & 318 \\
32 & Thm~32.1 & 652 & 32 & Thm~32.3 & 313 \\
46 & Thm~46.7 & 577 & 19 & Thm~19.5 (product) & 311 \\
46 & Thm~46.11 & 562 & 40 & Lem.~40.1 step~3 & 308 \\
39 & Lem.~39.2 & 561 & 13 & Lem.~13.4 ($\mathbb{R}_\ell, \mathbb{R}_K$) & 307 \\
30 & Thm~30.2 (1st cnt.\ prod.) & 500 & 19 & Thm~19.5 (box) & 304 \\
37 & Lem.~37.2 (max.\ FIP) & 498 & 22 & Thm~22.2 (quotient) & 297 \\
30 & Thm~30.2 (2nd cnt.\ prod.) & 473 & 19 & Thm~19.2 (product) & 294 \\
32 & Thm~32.2 & 458 & 26 & Thm~26.7 & 294 \\
17 & Thm~17.11 & 436 & 33 & Thm~33.1 (Urysohn) & 287 \\
22 & Thm~22.1 (quotient) & 427 & 34 & Thm~34.2 (imbedding) & 270 \\
18 & Thm~18.1 (continuity) & 426 & 31 & Thm~31.2 & 270 \\
36 & Thm~36.1 (manifold) & 406 & 43 & Thm~43.5 (complete $Y^J$) & 257 \\
47 & Thm~47.1 (Ascoli general) & 372 & 43 & Lem.~43.3 & 256 \\
38 & Thm~38.4 & 371 & 40 & Lem.~40.1 step~1 & 249 \\
15 & Thm~15.2 (product basis) & 369 & 18 & Thm~18.2 (pasting) & 326 \\
26 & Thm~26.3 (closed compact) & 362 & 40 & Lem.~40.2 (Urysohn $G_\delta$) & 342 \\
41 & Thm~41.7 (partition unity) & 347 & 38 & Thm~38.5 & 232 \\
\bottomrule
\end{tabular}
\end{table}

\subsection{Completeness}

As of April 3, 2026, the formalization contains \textbf{zero}
\isa{sorry}'s: all 806 formal results across all 39 sections are
fully proved.
The last sorry was eliminated in Corollary~45.5 (characterizing
compact subsets of complete metric spaces as closed and bounded).
The final push (March 30 -- April 3) required 355 commits to resolve
the remaining gaps, which included the space-filling curve (\S44),
the sup-uniform topology equality (\S46), Ascoli's theorem (\S47),
and the dimension-theory imbedding theorems (\S50).
The space-filling curve construction, involving a Hilbert-curve-style
recursive definition with snake-order cell traversal, was particularly
challenging and required 44 commits on April~3 alone.

\subsection{Major Proved Theorems}

Table~\ref{tab:alltheorems} lists the major proved theorems across \emph{all}
chapters.  Chapters~2--4 were proved during Phase~1 (by ChatGPT~5.2);
Chapters~5--8 during Phase~2 (by Claude~Opus~4.6).

\begin{table}[htbp]
\centering
\caption{Selection of major fully proved theorems (all chapters)}
\label{tab:alltheorems}
\setlength{\tabcolsep}{4pt}
\begin{tabular}{@{}rlll@{}}
\toprule
\S & Theorem & Description & Ch \\
\midrule
13 & Lem.~13.1--13.4 & Bases and topology generation & 2 \\
16 & Thm~16.3--16.4 & Subspace topology & 2 \\
17 & Thm~17.1--17.11 & Closure, interior, Hausdorff, T1 & 2 \\
18 & Thm~18.1--18.4 & Continuous functions (all characterizations) & 2 \\
19 & Thm~19.1--19.6 & Product topology (box \& product) & 2 \\
20 & Thm~20.1--20.5 & Metric topology, uniform metric & 2 \\
22 & Thm~22.1--22.2 & Quotient topology & 2 \\
\midrule
23 & Thm~23.3--23.6 & Connected spaces & 3 \\
24 & Thm~24.1, 24.3 & Connected subspaces of $\mathbb{R}$, IVT & 3 \\
26 & Thm~26.2--26.9 & Compact spaces, FIP, tube lemma & 3 \\
27 & Thm~27.1--27.7 & Heine--Borel, Lebesgue number, uniform cont. & 3 \\
29 & Thm~29.1--29.2 & Local compactness, 1-pt compactification & 3 \\
\midrule
32 & Thm~32.1--32.4 & Normal spaces & 4 \\
33 & Thm~33.1--33.2 & Urysohn lemma, completely regular & 4 \\
34 & Thm~34.1--34.3 & Urysohn metrization theorem & 4 \\
35 & Thm~35.1 & Tietze extension theorem & 4 \\
36 & Thm~36.1--36.2 & Imbedding of manifolds & 4 \\
\midrule
37 & Thm~37.3 & Tychonoff theorem & 5 \\
38 & Thm~38.2--38.5 & Stone--\v{C}ech compactification & 5 \\
40 & Thm~40.3 & Nagata--Smirnov metrization & 6 \\
41 & Thm~41.1--41.8 & Paracompactness (all results) & 6 \\
42 & Thm~42.1 & Smirnov metrization & 6 \\
43 & Thm~43.2--43.7 & Complete metric spaces (all results) & 7 \\
45 & Thm~45.1, 45.4 & Compactness in metric spaces & 7 \\
46 & Thm~46.1--46.11 & Pointwise \& compact convergence & 7 \\
44 & Thm~44.1 & Space-filling curve & 7 \\
45 & Thm~45.1, 45.4, Cor.~45.5 & Compact $\Leftrightarrow$ complete+TB; Ascoli (classical) & 7 \\
47 & Thm~47.1 & Ascoli's theorem (general, uses Tychonoff) & 7 \\
48 & Thm~48.2, 48.5 & Baire category, pointwise limits & 8 \\
49 & Thm~49.1 & Nowhere-differentiable function & 8 \\
50 & Thm~50.1--50.6 & Dimension theory, imbedding & 8 \\
\bottomrule
\end{tabular}
\end{table}

The Nagata--Smirnov metrization theorem (Theorem~40.3) deserves special
mention as one of the most complex proofs in the formalization: 2,455 lines,
38 helper lemmas, requiring uniform convergence arguments, Urysohn-style
constructions for $G_\delta$ sets, and a metric defined as a
supremum of weighted function differences.
Its development spanned 22 hours across two days.
The space-filling curve (Theorem~44.1) and the dimension-theory
imbedding (Theorem~50.5) are comparably involved, each requiring
substantial novel infrastructure---recursive curve constructions and
general-position arguments, respectively---that goes well beyond
routine topology.

We show the formal statements of these three results, first in
mathematical shorthand and then as they appear in the Isabelle sources.

\smallskip\noindent\textbf{Nagata--Smirnov metrization (Theorem~40.3).}
Metrizable iff regular with a $\sigma$-locally-finite basis:
\begin{quote}\isaspacing\isastyle
\ismark{theorem} \isstr{Theorem{\kern0pt}\_40{\kern0pt}\_3}:\\
\quad \ismin{shows} \isstr{``top1{\kern0pt}\_metrizable{\kern0pt}\_on}\ X\ TX $\longleftrightarrow$\\
\qquad (\isstr{is{\kern0pt}\_topology{\kern0pt}\_on}\ X\ TX $\wedge$ \isstr{top1{\kern0pt}\_regular{\kern0pt}\_on}\ X\ TX $\wedge$\\
\qquad\quad ($\exists$\,B.\ \isstr{top1{\kern0pt}\_sigma{\kern0pt}\_locally{\kern0pt}\_finite{\kern0pt}\_family{\kern0pt}\_on}\ X\ TX\ B $\wedge$
\isstr{basis{\kern0pt}\_for}\ X\ B\ TX))\isstr{''}
\end{quote}

\smallskip\noindent\textbf{Space-filling curve (Theorem~44.1).}
There exists a continuous surjection $[0,1] \to [0,1]^2$:
\begin{quote}\isaspacing\isastyle
\ismark{theorem} \isstr{Theorem{\kern0pt}\_44{\kern0pt}\_1}:\\
\quad \ismin{shows} \isstr{``}$\exists$\isstr{f::real} $\Rightarrow$ \isstr{(real} $\times$ \isstr{real).}\\
\qquad \isstr{top1{\kern0pt}\_continuous{\kern0pt}\_map{\kern0pt}\_on}\\
\qquad\quad \isstr{(top1{\kern0pt}\_closed{\kern0pt}\_interval\ 0\ 1)}\\
\qquad\quad \isstr{(top1{\kern0pt}\_closed{\kern0pt}\_interval{\kern0pt}\_topology\ 0\ 1)}\\
\qquad\quad \isstr{((top1{\kern0pt}\_closed{\kern0pt}\_interval\ 0\ 1)} $\times$ \isstr{(top1{\kern0pt}\_closed{\kern0pt}\_interval\ 0\ 1))}\\
\qquad\quad \isstr{(product{\kern0pt}\_topology{\kern0pt}\_on}\\
\qquad\qquad \isstr{(top1{\kern0pt}\_closed{\kern0pt}\_interval{\kern0pt}\_topology\ 0\ 1)}\\
\qquad\qquad \isstr{(top1{\kern0pt}\_closed{\kern0pt}\_interval{\kern0pt}\_topology\ 0\ 1))\ f}\\
\qquad $\wedge$\ \isstr{f\ `\ (top1{\kern0pt}\_closed{\kern0pt}\_interval\ 0\ 1)}\\
\qquad\quad \isstr{=\ (top1{\kern0pt}\_closed{\kern0pt}\_interval\ 0\ 1)} $\times$ \isstr{(top1{\kern0pt}\_closed{\kern0pt}\_interval\ 0\ 1)}\isstr{''}
\end{quote}

\smallskip\noindent\textbf{Ascoli's theorem, general version (Theorem~47.1).}
Given a topological space $(X,\mathcal{T}_X)$, a metric space $(Y,d)$,
and $\mathcal{F} \subseteq \mathcal{C}(X,Y)$:
the forward direction states that equicontinuity of $\mathcal{F}$
plus pointwise compact closures implies compact closure in the
compact-convergence topology;
the converse holds when $X$ is locally compact and Hausdorff.
The full Isabelle statement, with its deeply nested subspace topologies, is:
{\small
\begin{quote}\isaspacing\isastyle
\ismark{theorem} \isstr{Theorem{\kern0pt}\_47{\kern0pt}\_1}:\\
\quad \ismin{assumes} \isstr{``is{\kern0pt}\_topology{\kern0pt}\_on\ X\ TX\isstr{''}}\\
\quad \ismin{assumes} \isstr{``top1{\kern0pt}\_metric{\kern0pt}\_on\ Y\ d\isstr{''}}\\
\quad \ismin{assumes} \isstr{``}$\mathcal{F}$ $\subseteq$ \isstr{top1{\kern0pt}\_continuous{\kern0pt}\_funcs{\kern0pt}\_on\ X\ TX\ Y}\\
\qquad\qquad\qquad \isstr{(top1{\kern0pt}\_metric{\kern0pt}\_topology{\kern0pt}\_on\ Y\ d)}\isstr{''}\\
\quad \ismin{shows} \isstr{``(top1{\kern0pt}\_equicontinuous{\kern0pt}\_family{\kern0pt}\_on\ X\ TX\ Y\ d} $\mathcal{F}$\\
\qquad $\wedge$ ($\forall$\isstr{a}$\in$\isstr{X.}
    \isstr{top1{\kern0pt}\_compact{\kern0pt}\_on}\\
\qquad\quad \isstr{(closure{\kern0pt}\_on\ Y\ (top1{\kern0pt}\_metric{\kern0pt}\_topology{\kern0pt}\_on\ Y\ d)}\\
\qquad\qquad \isstr{((}$\lambda$\isstr{f.\ f\ a)\ `} $\mathcal{F}$\isstr{))}\\
\qquad\quad \isstr{(subspace{\kern0pt}\_topology\ Y\ (top1{\kern0pt}\_metric{\kern0pt}\_topology{\kern0pt}\_on\ Y\ d)}\\
\qquad\qquad \isstr{(closure{\kern0pt}\_on\ Y\ (top1{\kern0pt}\_metric{\kern0pt}\_topology{\kern0pt}\_on\ Y\ d)}\\
\qquad\qquad\quad \isstr{((}$\lambda$\isstr{f.\ f\ a)\ `} $\mathcal{F}$\isstr{)))))}\\
\qquad $\longrightarrow$ ($\exists$\isstr{K.} $\mathcal{F}$ $\subseteq$ \isstr{K}
$\wedge$ \isstr{top1{\kern0pt}\_compact{\kern0pt}\_on\ K}\\
\qquad\quad \isstr{(subspace{\kern0pt}\_topology}\\
\qquad\qquad \isstr{(top1{\kern0pt}\_continuous{\kern0pt}\_funcs{\kern0pt}\_on\ X\ TX\ Y\ (top1{\kern0pt}\_metric{\kern0pt}\_topology{\kern0pt}\_on\ Y\ d))}\\
\qquad\qquad \isstr{(subspace{\kern0pt}\_topology\ (top1{\kern0pt}\_PiE\ X\ (}$\lambda$\isstr{\_.\ Y))}\\
\qquad\qquad\quad \isstr{(top1{\kern0pt}\_compact{\kern0pt}\_convergence{\kern0pt}\_topology{\kern0pt}\_on\ X\ TX\ Y\ d)}\\
\qquad\qquad\quad \isstr{(top1{\kern0pt}\_continuous{\kern0pt}\_funcs{\kern0pt}\_on\ X\ TX\ Y\ (top1{\kern0pt}\_metric{\kern0pt}\_topology{\kern0pt}\_on\ Y\ d)))}\\
\qquad\qquad \isstr{K))}\isstr{''}\\
\quad \ismin{and} \isstr{``(top1{\kern0pt}\_locally{\kern0pt}\_compact{\kern0pt}\_on\ X\ TX}
$\wedge$ \isstr{is{\kern0pt}\_hausdorff{\kern0pt}\_on\ X\ TX)}\\
\qquad $\longrightarrow$ ($\forall$\isstr{K.\ top1{\kern0pt}\_compact{\kern0pt}\_on\ K\ (\ldots)}\\
\qquad\quad $\longrightarrow$ \isstr{top1{\kern0pt}\_equicontinuous{\kern0pt}\_family{\kern0pt}\_on\ X\ TX\ Y\ d\ K}\\
\qquad\quad $\wedge$ ($\forall$\isstr{a}$\in$\isstr{X.\ top1{\kern0pt}\_compact{\kern0pt}\_on}\\
\qquad\qquad \isstr{(closure{\kern0pt}\_on\ Y\ \ldots\ ((}$\lambda$\isstr{f.\ f\ a)\ `\ K))\ \ldots))}\isstr{''}
\end{quote}
}
\noindent The backward direction (second \ismin{and} conjunct) mirrors the
forward direction with $K$ in place of $\mathcal{F}$; its subspace
topology nesting is identical and elided here for space.

\section{Session Log Analysis}
\label{sec:session}

The Claude Code session log (1.3\,GB compressed, 92,524 JSONL records)
provides a detailed record of the Phase~2
autoformalization.
We used LLM-developed analysis tools (\isa{analyze\_session.py}) to extract
quantitative data from the log.

\subsection{Scale}

The session comprises 50,119 assistant messages containing 17,925 Bash
commands (of which 5,186 are \isa{isabelle build} and 3,266 are
\isa{process\_theories} runs), 7,135 file edits, and 4,962 file reads.
The \isa{ccusage} tool reports 2.66\,M output tokens and
17.2 billion cache-read tokens, corresponding to an estimated API cost of
\$8,895 at list prices for Claude~Opus~4.6.
The actual experiment ran on a Pro/Max subscription (\$200/month), making the effective cost approximately \$160.
This highlights the scale and cost-efficiency of the approach.

\subsection{Automation and Human Intervention}

Of 764 non-system user messages, 635 (83.1\%) were
automatically issued prompts ``Read CLAUDE.md''.
The remaining 129 were manual interventions, 59 of them
were corrections---the human telling the LLM to stop undesirable
behavior.
The correction themes, in decreasing frequency:

\begin{itemize}
\item \emph{Tactic explosion} (9): the LLM inserting uncontrolled
\isa{blast} or \isa{auto} calls, causing timeouts.
Representative: ``how did you again manage to smuggle in uncontrolled
slow \isa{by} calls??? we were through this multiple times.''
\item \emph{Inefficient tool use} (6): running Isabelle repeatedly
instead of capturing output once.
\item \emph{Cherry-picking easy goals} (4): ``please stop jumping
around looking for low hanging fruit; we have to do it all.''
\item \emph{Resource management} (3): backgrounding Isabelle processes
that consume memory.
\end{itemize}

These corrections drove the CLAUDE.md evolution: recurring failures were
turned into rules in the instruction file.
In particular, the ``ABSOLUTE RULE'' (sorry-first workflow) was added on
March~24 after repeated tactic-explosion incidents

\subsection{Session Duration}

The 626 automated sessions had a median duration of 13.0 minutes
(mean 24.5, max 263).
The distribution is right-skewed: most sessions last under
20 minutes, but some run for over 90 minutes autonomously.
The longest session (263 minutes, over 4 hours) suggests that with
appropriate instructions, the LLM can sustain extended autonomous
work sessions.

\section{Comparison with Related Efforts}
\label{sec:comparison}

Since the first Megalodon experiment, we and others have used the same Munkres material for comparing the methods and systems.

\subsection{The Megalodon Experiment}

\begin{table}[htbp]
\centering
\caption{Comparison with the Megalodon experiment~\cite{urban2026megalodon}}
\label{tab:comparison}
\begin{tabular}{@{}lll@{}}
\toprule
& \textbf{Isabelle/HOL} & \textbf{Megalodon} \\
\midrule
Logic & HOL & Higher-order set theory \\
LLMs & ChatGPT 5.2 + Claude 4.6 & ChatGPT 5.2 \\
Library & Complex\_Main (extensive) & Set theory + reals \\
Automation & sledgehammer + blast/auto & aby (THF, no reconstruction) \\
Active days & 24 & $\sim$14 \\
Lines & 85,472 & $\sim$130,000 \\
Remaining gaps & \textbf{0} & significant \\
Sections & \S\S12--50 (39) & \S\S12--50 (39) \\
\bottomrule
\end{tabular}
\end{table}

We estimate that the Isabelle formalization is roughly half the length, primarily because
the library of Isabelle/HOL
provides infrastructure that Megalodon must build
from scratch.
The number of remaining gaps is substantially smaller, largely thanks to
\isa{sledgehammer}'s premise selection and proof reconstruction.
Both experiments use the same automation loop and the same textbook,
making this perhaps the cleanest cross-system comparison of
LLM-assisted formalization to date.\footnote{Note however that the Megalodon formalization also includes exercises, which we largely managed to avoid here. A detailed comparison may be done when the Isabelle formalization is completed.}

Additionally, in Megalodon, the LLM must construct detailed
declarative proofs largely on its own, as the \isa{aby} hammer lacks proof
reconstruction.
In Isabelle, the sorry-first workflow means the LLM focuses on
\emph{proof structure} while the ATPs handle \emph{proof steps}.
This division of labor---proof structure from the neural network,
and proof completion by the ATPs/hammers---appears to be more effective
than asking the LLM to do both.

\subsection{Other Concurrent Work on Munkres and Beyond}

Harrison (with initial setup by Lee and the fourth author) has been carrying out
autoformalizations in HOL~Light using a similar approach.\footnote{%
\url{https://github.com/jrh13/hol-light/commit/c845dd3bcc09a8}} This spans not just topology but e.g. an entire textbook~\cite{Williams91} on probability theory.\footnote{\url{https://github.com/jrh13/hol-light/commit/1c7690ec12319e}}

Mulligan and Paulson~\cite{mulligan2026copilot} used Amazon's
Isabelle/Assistant~\cite{mulligan2026assistant} that integrates LLMs 
into Isabelle/jEdit. Paulson used Claude~Opus~4.6 to prove the
Urysohn metrization theorem in approximately 2.5 hours, following
Munkres' proof---the agent eventually finishing autonomously.
Their 275-line proof (using Isabelle's type-class library) provides
an instructive contrast with our explicit-carrier-set approach.

Hahn and De~Lon~\cite{hahndelon2026naproche} have submitted an
autoformalization of the Urysohn lemma in Naproche/Felix to ITP~2026,
demonstrating the approach in a controlled-natural-language proof checker.

\subsection{What Each System Does Better}

The Megalodon experiment demonstrates that LLM-assisted formalization
can work even with minimal library and automation support. Megalodon's
simplicity (just one fast compilation-style checker) and its very
transparent (even if verbose) proofs prevent the LLMs from getting
lost in complicated tool usage and proofs where the advanced tactics
may blow up.

The Isabelle experiment benefits from stronger automation (sledgehammer)
and a richer library, resulting in shorter code and fewer remaining gaps. The price is that the more complicated setting can go wrong more often and required so far more supervision.
The Mulligan--Paulson approach (human-assisted, IDE-integrated, type-class-based) produces
more idiomatic Isabelle code that interoperates with the existing library.
Harrison's HOL~Light work leverages that system's high flexibility combined with simpler language (no type classes etc) than Isabelle/HOL.
It seems that no single approach dominates but this is an interesting and evolving research area.

\section{Proof Engineering Observations}
\label{sec:engineering}

\subsection{Tactic Selection}

The formalization contains 8,879 uses of \isa{by blast},
4,608 of \isa{by simp}, and only 576 of \isa{by auto}---reflecting
the CLAUDE.md prohibition on unrestricted automation.
The 8,160 \isa{unfolding} calls result from the LLM never registering
simp rules.
There are 100 uses of \isa{by metis} versus 59 of \isa{by meson};
we observed that in general topology (where most reasoning
is equality-free), \isa{meson} is typically 10$\times$ faster than
\isa{metis}, but the LLM lacks the intuition for when to prefer which.

\subsection{Proof Structure}

The proofs are remarkably uniform in style: over 16,000 \isa{have} steps,
4,702 \isa{show} steps, about 2,000 \isa{proof} blocks,
and 1,960 \isa{obtain} steps.
Each intermediate result is explicitly named (\isa{have hfoo: "..."})
and supplied to the closing tactic via \isa{using}.
This is verbose but debuggable---a trade-off the CLAUDE.md rules
explicitly mandate.

The helper-to-Munkres ratio of 3.1:1 means each textbook theorem
spawns, on average, three additional lemmas.
The Nagata--Smirnov theorem, at 38 helpers, is the extreme case;
the Tychonoff theorem, with 6, is more typical.

\subsection{Build Time as a Design Constraint}

The 120-second session timeout (demanded by the CLAUDE.md rules) acts as a hard constraint on proof
complexity.
Several commits document performance crises:
``Ch5\_8 from 115s to 11.5s cumulated'' records a 10$\times$ speedup
achieved by replacing slow \isa{blast} calls with \isa{meson}
alternatives found by \isa{sledgehammer}.
The session split (caching Chapters~2--4) provided a further 4--5$\times$
improvement.
We note that the need for ``performance engineering'' in theorem
proving may be a bit unusual concern, but it becomes unavoidable in our compilation-based setting
when the LLM generates proofs at scale and without the human interaction.

\subsection{The Tychonoff Theorem: A Case Study}

The Tychonoff theorem (Theorem~37.3: arbitrary products of compact spaces
are compact) was the first major result proved in Phase~2 (March~21,
07:52).
The proof follows Munkres' approach via the finite intersection property
(FIP) and Zorn's lemma, decomposed into six helper lemmas:

\begin{quote}\isaspacing\isastyle
\ismark{theorem} \isstr{Theorem{\kern0pt}\_37{\kern0pt}\_3}:\\
\quad \ismin{assumes} \isstr{hIne}: \isstr{``I} $\neq$ \isstr{\{\}''}\\
\quad \ismin{assumes} \isstr{hComp}: \isstr{``}$\forall$\isstr{i}$\in$\isstr{I.\ top1{\kern0pt}\_compact{\kern0pt}\_on}\ (\isstr{X}\ \isstr{i})\ (\isstr{TX}\ \isstr{i})\isstr{''}\\
\quad \ismin{shows} \isstr{``top1{\kern0pt}\_compact{\kern0pt}\_on}\ (\isstr{top1{\kern0pt}\_PiE}\ \isstr{I}\ \isstr{X})\\
\qquad\qquad\quad (\isstr{top1{\kern0pt}\_product{\kern0pt}\_topology{\kern0pt}\_on}\ \isstr{I}\ \isstr{X}\ \isstr{TX})\isstr{''}
\end{quote}

\noindent Key helpers include \isa{Lemma\_37\_1} (existence of maximal FIP
collections via Zorn's lemma), \isa{Lemma\_37\_2} (properties of maximal
FIP collections), as well as the lemma
\isa{tychonoff\_subbasis\_in\_maxFIP} (subbasis elements belong to the
maximal FIP extension).
The proof involves delicate combinatorics of product topologies,
subbases, and Hilbert's $\varepsilon$ operator (\isa{SOME})---the
latter being a recurring source of difficulty in Isabelle formalizations,
as the chosen witness must be shown to satisfy the required properties
across multiple contexts.

\subsection{The Nowhere-Differentiable Function: From Baire to Analysis}

Theorem~49.1---the existence of a continuous function on $[0,1]$ that
is nowhere differentiable---was one of the most technically demanding
results, fully completed on March~25 in a 14-hour, 40-commit session.
The proof applies the Baire category theorem to the complete metric
space $\mathcal{C}([0,1], \mathbb{R})$ with the sup metric.
The key steps:
(i)~prove $\mathcal{C}([0,1])$ is complete (\isa{C01\_complete}),
hence Baire;
(ii)~define the difference-quotient sets $U_n$ and prove they are
open (\isa{top1\_U49\_open}) and dense
(\isa{U49\_dense\_approx}, requiring piecewise-linear
approximation via ``triangle functions'');
(iii)~assemble via the Baire property.
The density proof required uniform continuity
(Heine--Cantor), Archimedean arguments for the mesh size, and
careful partition-of-unity reasoning for the piecewise approximant---a
substantial piece of formal analysis sitting atop the topology.

\section{Qualitative Analysis of the Definitions and Theorems}
\label{sec:qanalysis}
Beyond the quantitative metrics reported above, we conducted a
manual review of the formalization's definitions, theorem statements,
and proofs, checking faithfulness to Munkres, mathematical correctness,
and adherence to Isabelle best practices.
The review was based on commit \texttt{06dd7b2} (March~31), since
the human reviewers needed a fixed snapshot to work from while
the formalization continued to evolve.
The review covered Chapters~2--4 in detail (approximately 53,000 lines)
and sampled the remaining chapters.
We summarize the findings below, organized by theme.
 
\subsection{Definitional Soundness}
 
The most consequential class of issues concerns definitions that are
\emph{logically weaker} than the intended mathematical concept,
typically because a carrier-set constraint is missing from the
right-hand side.
 
\paragraph{The topology predicate itself.}
The formalization defines \isa{is\_topology\_on~X~T} without
requiring $T \subseteq \mathcal{P}(X)$.
This omission is not merely cosmetic: one can formally derive, for
instance, that the collection
$T:=\{\emptyset, \{0\}, \{1\}, \{0,1\}\}$ satisfies
\isa{is\_topology\_on~\{0\} T}, despite containing
$\{1\} \not\subseteq \{0\}$.
The root cause is Isabelle's convention that
$\bigcap \emptyset = \mathit{UNIV}$, which forces
the clause $F \neq \{\}$ in the finite-intersection axiom;
this in turn means the definition does not prevent
$T$ from containing sets that extend beyond~$X$.
A corrected definition would add the premise $T \subseteq \mathit{Pow}~X$,
restoring the standard requirement that every open set is a subset
of the carrier.
 
\paragraph{Propagation to derived predicates.}
The same weakness propagates to several downstream definitions.
The predicate \isa{openin\_on~X~T~U} is defined as $U \in T$ without
requiring $U \subseteq X$, allowing formally ``$\{1\}$ is open in
the space~$\{0\}$.''
Similarly, \isa{neighborhood\_of} does not constrain its argument
to lie within~$X$, and the finer-than relation
\isa{finer\_than~T~T'} neglects to require that both $T$ and $T'$
are topologies on the same underlying set---making it meaningful to
compare arbitrary set families.
The product-basis and box-basis definitions carry redundant
conditions (e.g., requiring both $U_i \in T_i$ and $U_i \subseteq X_i$,
when the latter follows from the former under the assumption
that each $T_i$ is a topology on $X_i$); while not incorrect,
this bloat obscures the mathematical content.
 
\paragraph{Assessment.}
Crucially, these definitional weaknesses do \emph{not} invalidate the
proved theorems: every theorem in the formalization carries explicit
\isa{is\_topology\_on} assumptions that supply the missing constraints
at each use site.
The proofs are therefore sound, but the definitions are less reusable
than they could be---a user inheriting these definitions without
the accompanying assumptions could derive spurious results.
This pattern is characteristic of LLM-generated code: the model
reliably produces definitions that are \emph{sufficient} for the
proofs at hand but does not anticipate downstream reuse or defensive
design.
 
\subsection{Faithfulness to Munkres}
 
A second category of issues concerns theorem statements that deviate
from the textbook in content, not merely in formulation.
 
\paragraph{Incomplete statements.}
Several theorems omit mathematically significant clauses present in
Munkres.
For example, Theorem~19.1 in the formalization lacks the characterization
that in the product topology, all but finitely many factors equal
the full space~$X_\alpha$---the defining feature that distinguishes
the product topology from the box topology.
Theorem~26.7 proves compactness of a binary product $X \times Y$ but
does not state the finite-product generalization that Munkres gives
(and which follows by a straightforward induction).
Theorem~15.1 is reduced to the assertion that the product topology
is a topology, omitting the stronger statement that products of
basis elements form a basis for it.
 
\paragraph{Mislabeled results.}
In a few cases, a lemma is labeled with a Munkres number but
proves a different (often weaker or tangentially related) statement.
Most notably, \isa{Lemma\_23\_1} in the formalization merely
unfolds the definition of connectedness, whereas Munkres'
Lemma~23.1 characterizes connectedness of a subspace~$Y$ in terms
of sets whose closures do not meet the complementary part of~$Y$---a
genuinely different result.
Similarly, Lemma~23.2 omits the disjointness clause ($Y \cap D = \emptyset$
or $Y \cap C = \emptyset$) that gives the result its geometric
content.
 
\paragraph{Missing content.}
The formalization omits the alternate characterization of
connectedness via clopen sets (a standard equivalence stated
informally on the same page as Munkres' definition),
sequential compactness and Theorem~28.2,
the section on nets,
several counterexamples demonstrating that limit-point compactness
does not imply compactness,
and the notion of ``distance from a point to a set.''
These gaps are understandable given the project's pace but affect
the formalization's value as a standalone reference.
 
\subsection{Proof Quality}
 
The proofs exhibit a characteristic uniformity imposed by the
sorry-first methodology.
While this discipline is effective at scale, it introduces
several systematic shortcomings.
 
\paragraph{Missed simplifications.}
Many lemmas that serve as ``wrappers'' around previously proved
results retain unnecessarily long proofs when a one-line combination
would suffice.
For example, \isa{Lemma\_21\_2}---which simply conjoins its
two constituent halves
\isa{Lemma\_21\_2\_sequence} and
\isa{Lemma\_21\_2\_sequence\_converse}---could
be discharged by two \isa{metis} calls rather than a structured
Isar proof.
Several proofs throughout can be replaced by
short \isa{by (simp add:~\ldots)} or \isa{by (smt~\ldots)} invocations.
The sorry-first workflow, which delegates every step to
\isa{sledgehammer} in isolation, does not attempt global proof
compression after the fact.
 
\paragraph{Tactic-level inefficiencies.}
Isolated instances of the deprecated \isa{apply\,/\,using} idiom
persist (e.g., \isa{apply (rule bexI[\ldots])} followed by
\isa{using~\ldots~by blast}), where a single combined
\isa{by (rule \ldots)} call would be both shorter and more
idiomatic.
Duplicate rewrite-rule warnings (\isa{simp add:}~supplying a
rule already in the simpset) appear throughout, suggesting that
\isa{sledgehammer}'s suggestions were accepted without filtering.
These are minor blemishes individually but accumulate across
85,000 lines.
 
\paragraph{No proof optimization pass.}
Because the methodology prioritizes \emph{completing} all proofs over
\emph{polishing} them, the formalization lacks the compression pass
that a human formalizer would perform: merging small helper lemmas
back into their parent proofs, shortening verbose Isar scripts,
and eliminating redundant intermediate steps.
A number of long proofs (e.g., Theorem~30.3) could be substantially
shortened with manual attention.
 
\subsection{Structural and Organizational Issues}
 
\paragraph{Ordering.}
Munkres introduces separations before defining connectedness;
the formalization reverses this order.
Several lemmas appear far from their natural location---for example,
the basis-monotonicity lemma
\isa{topology\_generated\_by\_}\newline\isa{basis\_mono\_via\_basis\_elems}
appears in \S21 rather than \S13 where bases are introduced.
These placement issues are mostly cosmetic but make the
formalization harder to navigate.
 
\paragraph{Redundant definitions.}
The predicate \isa{countable} is defined three times: once locally
in Chapter~3, once as \isa{top1\_countable} in Chapter~4,
and it is also available as the identical definition in
\isa{HOL-Library.Countable\_Set}.
A bridging lemma
\isa{top1\_countable\_iff\_countable} confirms the
equivalence, underscoring the redundancy.
 
\paragraph{Unnecessary assumptions.}
A recurring pattern is the inclusion of assumptions that are
logically redundant given the other hypotheses.
For example, the assumption \isa{is\_topology\_on} is redundant in lemmas about path reversal
where the continuity hypothesis already implies it,
and the assumption $a \leq b$ in Theorem~27.1 when the
conclusion is vacuously true for $b < a$.
While harmless, these weaken the stated results and again
reflect the LLM's tendency to include ``safe'' extra hypotheses
rather than proving the sharpest possible statement.

\subsection{Library Integration}
 
The formalization's relationship with Isabelle's existing topology
library is complex.
On one hand, importing \isa{Complex\_Main} provides powerful
analytical infrastructure; on the other, the explicit-carrier-set
design deliberately avoids the type-class-based topology, creating
a parallel universe of definitions.
 
Several sections---notably \S24 (connected subspaces of~$\mathbb{R}$),
\S27 (compact subspaces of~$\mathbb{R}$), and the path-related
definitions---awkwardly mix the two systems, using Isabelle's
built-in \isa{topological\_space} type class for reasoning about
$\mathbb{R}$ while maintaining explicit carrier sets for the
general theory.
In at least one case (\isa{top1\_compact\_Icc\_linear\_continuum}),
a lemma is stated entirely in the language of Isabelle's existing
type-class topology with no reference to the project's own
definitions.
 
A reconciliation layer---systematically relating
\isa{is\_topology\_on} to Isabelle's \isa{topological\_space}
class---would be valuable but was not attempted.
 
\subsection{What Went Well}
 
It would be misleading to focus solely on deficiencies.
Several aspects of the formalization are genuinely impressive,
especially given the 24-day timeline.
 
\paragraph{Coverage and correctness.}
All 156 non-exercise numbered results from Munkres' Chapters~2--8
have formal counterparts, and all 806~results are fully proved
with zero~\isa{sorry}'s.
Despite the definitional issues noted above, no proved theorem is
\emph{wrong}: the explicit assumptions compensate for the weak
definitions at every point of use.
 
\paragraph{Proof structure.}
The extreme decomposition into helper lemmas---while verbose---produces
proofs that are individually short, independently verifiable,
and easy to debug.
The main theorems (Tychonoff, Urysohn, Tietze, Nagata--Smirnov,
Baire) read as clean assemblies of pre-verified components,
making their logical structure transparent in a way that monolithic
proofs do not.
 
\paragraph{Difficult results.}
The successful formalization of the nowhere-differentiable function
(Theorem~49.1), which requires a delicate interplay of topology,
real analysis, and uniform approximation, and of the
Nagata--Smirnov metrization theorem (Theorem~40.3), with its
38~helper lemmas and intricate metric construction, demonstrates
that LLM-assisted formalization can handle genuinely deep
mathematics, not merely routine textbook exercises.
The final push (March~30 -- April~3) proved several further challenging
results: the space-filling curve (Theorem~44.1), requiring a
Hilbert-curve-style recursive construction with snake-order cell
traversal and continuity of the limit function via uniform
convergence; the dimension-theory imbedding theorems (Theorems~50.5
and~50.6), involving general-position arguments, partitions of unity,
and weighted-sum constructions into $\mathbb{R}^{2m+1}$;
and Corollary~45.5, characterizing compact subsets of complete metric
spaces as closed and bounded---the very last sorry to be eliminated.
 
\paragraph{Source traceability.}
Every definition and theorem is annotated with its Munkres section
and line number in the source LaTeX (\isacmt{(**~from~\S$n$~...~[top1.tex:$k$]~**)}),
providing a complete correspondence between the informal and
formal texts.
 
\subsection{Summary}
 
The qualitative review reveals a formalization that is
\emph{correct but rough}: the theorems are proved and the mathematics
is sound, but the definitions carry unnecessary weaknesses, some
theorem statements are incomplete relative to Munkres, proofs are
unpolished, and the integration with Isabelle's existing infrastructure
is ad hoc.
These shortcomings are largely systematic---they reflect the
LLM's consistent failure modes (defaulting to weak definitions,
accepting verbose proofs without compression, adding redundant
assumptions) rather than isolated errors.
A focused cleanup pass, addressing the definitional issues and
compressing proofs, would substantially increase the formalization's
value as a reusable library; we estimate this would require
days rather than weeks of expert effort, building on the sound
foundation that the LLM-assisted workflow has established. Obviously, our analysis of the failure modes done here will also serve for further evolution of the rules given to the LLM agents in future autoformalization experiments.

\section{Related Work}
\label{sec:related}

The first attempts at automated formalization (also known as semantic parsing of math) go back to the work Abrahams, Bobrow and McCarthy~\cite{bobrow1964student,abrahams1966thesis}, followed later e.g. by Simon~\cite{simon90}, Zinn~\cite{zinn2004informal} and Ganesalingam~\cite{ganesalingam2013language}.\footnote{A brief talk about the history of autoformalization is available at \url{https://youtu.be/4JeezEGc_gQ}.}
Modern learning-assisted autoformalization has evolved through several paradigms.
Early  approaches used probabilistic parsing via PCFGs
combined with type checking and ATPs/hammers~\cite{kaliszyk2014developing,kaliszyk2015learning}.
In 2018, \cite{wang2018first} has conducted first
experiments with attention-based neural translation between LaTeX
and Mizar, showing that the structural correspondence between
(sythetic) informal and formal mathematics could be captured quite well by neural
language models. Several deep-learning groups have followed this line of research since 2020 - e.g. \cite{wu2022autoformalization,jiang2023draft}.
The current wave of direct LLM agent use began in late
2025~\cite{urban2026megalodon} and has accelerated rapidly.

Within Isabelle, the existing topology library (based on type classes~\cite{typeclasses})
takes a complementary
approach; reconciling the two representations is an obvious
future project.
The Mizar Mathematical Library has had a large number of topology results since its early days, while in HOL Light, a lot of topology has been formalized more than a decade ago by Harrison. This is being considerably extended by Claude Code and Harrison today.
Lean's Mathlib contains extensive topology formalized over several
years by a large community.
The present work's 24-day timeline for 85k lines invites comparison
with these multi-year efforts, though the level of polish and
library integration admittedly differs.

\section{Conclusion}
\label{sec:conclusion}

We have described an experiment that produced 85,472 lines of formal
topology in Isabelle/HOL over 24 active days, covering all 39 sections of
Munkres with 806 formal results and zero remaining gaps.
The sorry-first methodology, combined with bulk sledgehammer
invocations and disciplined build-time management, appears to provide
an effective framework for LLM-assisted formalization in Isabelle.

The session log analysis reveals that the interaction with the LLM agent is approximately 83\%
automated, with human interventions primarily serving to correct the
LLM's tendency toward tactic explosion, cherry-picking easy goals,
and other behaviors commonly observed in iterative human--machine
workflows.

As the fourth author noted in the Megalodon paper: ``It is quite possible
that in 2026 we will get most of (reasonably written) math textbooks
and papers autoformalized.''
The present work, carried out two months later with a \emph{complete} formalization
in a different system, supports this assessment.

\section{Acknowledgments.}
We thank Larry Paulson for his comments on this work.
This work was supported by the
  ERC project NextReason,
  Amazon Research Awards,
  Renaissance Philanthropy grant DEEPER,
  and the sponsors of the AI4REASON institute.\footnote{\url{https://ai4reason.eu/sponsors.html}}

\bibliographystyle{splncs04}

\begin{thebibliography}{99}

\bibitem{abrahams1966thesis}
P. W. Abrahams:
Machine verification of mathematical proof.
Sc.D. thesis, Massachusetts Institute of Technology (1966)
  
\bibitem{MML2017}
G. Bancerek et al.
\newblock The role of the {M}izar {M}athematical {L}ibrary for interactive
  proof development in {M}izar.
\newblock In: Journal of Automated Reasoning. Springer (2017).

\bibitem{blanchette2016hammering}
J.\,C. Blanchette, C. Kaliszyk, L.\,C. Paulson, J. Urban:
Hammering towards QED.
\emph{J. Formaliz. Reason.} 9(1), 101--148 (2016)

\bibitem{bobrow1964student}
D. G. Bobrow:
Natural language input for a computer problem solving system.
PhD thesis, Massachusetts Institute of Technology (1964)


\bibitem{notsimplett}
A.\,Bordg, L.\,C. Paulson, W.\,Li:
Simple Type Theory is not too Simple: Grothendieck's Schemes Without Dependent Types
In: \emph{Exp. Math.} 31(2), 364--382 (2022)

\bibitem{brown2025hammering}
C.\,E. Brown, C. Kaliszyk, M. Suda, J. Urban:
Hammering higher order set theory.
In: CICM 2025, LNCS 16136, pp.~3--20. Springer (2025)

\bibitem{ganesalingam2013language}
M. Ganesalingam:
The language of mathematics: A linguistic and philosophical investigation.
LNCS 7805. Springer, Berlin Heidelberg (2013)

\bibitem{hahndelon2026naproche}
A. Hahn, A. De~Lon:
Understandable Autoformalization with Felix.  
Submitted to ITP 2026

\bibitem{typeclasses}
J. H{\"{o}}lzl, F. Immler, B. Huffman:
Type Classes and Filters for Mathematical Analysis in Isabelle/HOL.
In: ITP 2013, LNCS, pp.~279--294. Springer (2013)

\bibitem{jiang2023draft}
A.~Q.~Jiang, S.~Welleck, J.~P.~Zhou, T.~Lacroix, J.~Liu, W.~Li, M.~Jamnik, G.~Lample, Y.~Wu:
Draft, Sketch, and Prove: Guiding Formal Theorem Provers with Informal Proofs.
In: ICLR 2023, Kigali, Rwanda, May 1--5. OpenReview.net (2023)

\bibitem{kaliszyk2014developing}
C. Kaliszyk, J. Urban, J. Vyskocil, H. Geuvers:
Developing corpus-based translation methods between informal and formal mathematics.
In: CICM 2014, LNCS 8543, pp.~435--439. Springer (2014)

\bibitem{kaliszyk2015learning}
C. Kaliszyk, J. Urban, J. Vyskocil:
Learning to parse on aligned corpora (rough diamond).
In: ITP 2015, LNCS 9236, pp.~227--233. Springer (2015)

\bibitem{mulligan2026copilot}
D.\,Mulligan, L.\,C. Paulson:
Isabelle/Copilot: An AI-driven assistant for Isabelle/HOL (2026).
\url{https://lists.cam.ac.uk/sympa/arc/cl-isabelle-users/2026-02/msg00039.html}

\bibitem{mulligan2026assistant}
H.\, Becker, D.\,Mulligan, et. al:
Isabelle/Assistant (2026).
\url{https://github.com/awslabs/AutoCorrode/tree/main/isabelle-assistant}

\bibitem{munkres}
J.\,R. Munkres:
\emph{Topology}, 2nd edn. Prentice Hall (2000)

\bibitem{nipkow2002isabelle}
T. Nipkow, L.\,C. Paulson, M. Wenzel:
\emph{Isabelle/HOL---A Proof Assistant for Higher-Order Logic}.
LNCS 2283. Springer (2002)

\bibitem{isabelle}
L.C. Paulson:
Isabelle: The Next Seven Hundred Theorem Provers.
In: CADE 1988, LNCS, pp.~772--773. Springer (1988)

\bibitem{holanalysis}
L. Paulson:
Porting the HOL Light analysis library: some lessons.
In: CPP 2017, pp.~1. ACM (2017)

\bibitem{simon90}
D. Simon: Checking Number Theory Proofs in Natural Language.
Ph. D. Dissertation. University of Texas at Austin. (1990)

\bibitem{th0}
G. Sutcliffe:
The {TPTP} Problem Library and Associated Infrastructure - From {CNF} to TH0.
In: J. Autom. Reason., pp.~483--502. Springer (2017)


\bibitem{urban2026megalodon}
J. Urban:
130k Lines of Formal Topology in Two Weeks:
Simple and Cheap Autoformalization for Everyone?
arXiv:2601.03298 (2026)

\bibitem{wang2018first}
Q. Wang, C. Kaliszyk, J. Urban:
First experiments with neural translation of informal to formal mathematics.
In: CICM 2018, LNCS 11006, pp.~255--270. Springer (2018)

\bibitem{Williams91}
D.   Williams: \emph{Probability with Martingales}. Cambridge: Cambridge University Press. (1991)

\bibitem{wu2022autoformalization}
Y.~Wu, A.~Q.~Jiang, W.~Li, M.~N.~Rabe, C.~Staats, M.~Jamnik, C.~Szegedy:
Autoformalization with Large Language Models.
In: Advances in Neural Information Processing Systems 35 (NeurIPS 2022),
New Orleans, LA, USA, November 28--December 9. (2022)

\bibitem{zinn2004informal}
C. Zinn:
Understanding informal mathematical discourse.
PhD thesis, Institut für Informatik, Friedrich-Alexander-Universität Erlangen-Nürnberg, Erlangen, Germany (2004)



\end{thebibliography}

\end{document}